\documentclass[3p,11pt]{elsarticle}

\usepackage{amsmath,amssymb,bm}
\usepackage{siunitx,chemformula}
\usepackage{url}
\usepackage[colorlinks=true,
            linkcolor=blue,
            citecolor=blue,
            urlcolor=blue]{hyperref}

\usepackage[normalem]{ulem}
\usepackage[table]{xcolor}
\usepackage{listings}
\usepackage{booktabs}
\DeclareMathOperator{\E}{\mathbb{E}}
\DeclareMathOperator*{\argmin}{\arg\,\min}

\definecolor{codegreen}{rgb}{0,0.6,0}
\definecolor{codegray}{rgb}{0.5,0.5,0.5}
\definecolor{codepurple}{rgb}{0.58,0,0.82}
\definecolor{backcolour}{rgb}{0.95,0.95,0.92}

\lstdefinestyle{mystyle}{
    backgroundcolor=\color{backcolour},
    commentstyle=\color{codegreen},
    keywordstyle=\color{magenta},
    numberstyle=\tiny\color{codegray},
    stringstyle=\color{codepurple},
    basicstyle=\ttfamily\footnotesize,
    breaklines=true,
    numbers=left,
    numbersep=5pt,
    showstringspaces=false,
    tabsize=2
}
\usepackage{tgheros}

\usepackage{fancyhdr}
\begin{document}

\begin{frontmatter}

\title{\textbf{An AI-driven framework for the prediction of personalised\\health response to air pollution}}

\author[DSI]{Nazanin Zounemat-Kermani\textsuperscript{\dag}}
\author[ESE]{Sadjad Naderi\textsuperscript{\dag}}
\author[ESE]{Claire H.~Dilliway}
\author[ESE,IX]{Claire E.~Heaney\textsuperscript{*}}
\author[ESE]{Shrreya Behll}
\author[ESE]{Boyang Chen}
\author[NHLI]{Hisham Abubakar-Waziri}
\author[Mat]{Alexandra E. Porter}
\author[SPH]{Marc Chadeau-Hyam}
\author[DSI]{Fangxin Fang}
\author[NHLI]{Ian M. Adcock}
\author[DSI,NHLI]{Kian Fan Chung}
\author[DSI,ESE,IX]{Christopher C. Pain}

\affiliation[DSI]{
  organization={Data Science Institute, Imperial College London},
  country={UK}
}

\affiliation[ESE]{
  organization={Department of Earth Science \& Engineering, Imperial College London},
  country={UK}
}

\affiliation[IX]{
  organization={Centre for AI-Physics Modelling, Imperial-X, Imperial College London},
  country={UK}
}

\affiliation[NHLI]{
  organization={National Heart \& Lung Institute, Imperial College London},
  country={UK}
}

\affiliation[Mat]{
  organization={Department of Materials, Imperial College London},
  country={UK}
}

\affiliation[SPH]{
  organization={MRC/PHE Centre for Environment and Health, School of Public Health, Imperial College London},
  country={UK}
}

\begin{abstract}
Air pollution is a growing global health threat, exacerbated by climate change and linked to cardiovascular and respiratory diseases. While personal sensing devices enable real-time physiological monitoring, their integration with environmental data for individualised health prediction remains underdeveloped. Here, we present a modular, cloud-based framework that predicts personalised physiological responses to pollution by combining wearable-derived data with real-time environmental exposures. At its core is an Adversarial Autoencoder (AAE), initially trained on high-resolution pollution-health data from the INHALE study and fine-tuned using smartwatch data via transfer learning to capture individual-specific patterns. Consistent with changes in pollution levels commonly observed in the real-world, simulated pollution spikes (+100\%) revealed modest but measurable increases in vital signs (e.g., +2.5\% heart rate, +3.5\% breathing rate). To assess clinical relevance, we analysed U-BIOPRED data and found that individuals with such subclinical vital sign elevations had higher asthma burden scores or elevated Fractional Exhaled Nitric Oxide (FeNO), supporting the physiological validity of these AI-predicted responses. This integrative approach demonstrates the feasibility of anticipatory, personalised health modelling in response to environmental challenges, offering a scalable and secure infrastructure for AI-driven environmental health monitoring.
\end{abstract}


\end{frontmatter}

\renewcommand{\thefootnote}{\fnsymbol{footnote}}
\footnotetext{\textsuperscript{*}Corresponding author Claire E.~Heaney \texttt{\small c.heaney@imperial.ac.uk}.}
\footnotetext{\textsuperscript{\dag}Denotes shared first authorship.}

\section{Introduction}
Air pollution poses a significant threat to public health, with over 90\% of the global population being exposed to levels in excess of WHO guidelines~\citep{WHO}. Each year, air pollution contributes to approximately 7 million premature deaths worldwide, primarily through cardiovascular and respiratory diseases such as ischemic heart disease, stroke, lung cancer, chronic obstructive pulmonary disease (COPD), asthma and other respiratory infections~\citep{Yu,Guan,Lee}. In the UK, air pollution is recognised as a severe public health issue, comparable to cancer, obesity and heart disease~\citep{PHE2019}. The annual mortality of human-made air pollution in the UK is estimated to be between 28,000 and 36,000 deaths every year~\citep{PHE2019}. Despite efforts to reduce pollution levels, they remain high, and therefore there is a critical need to protect individuals by minimising their personal exposure, thus mitigating the adverse health effects of pollution~\citep{PinhoGomes}. 

Other  environmental factors such as urban design, temperature, humidity, and wind patterns also influence the onset and progression of chronic respiratory diseases. A recent exposome study across 14 European cities showed that people exposed to both air pollution and limited green space had a higher risk of developing asthma~\citep{Lee}. In addition, COPD risk factors include indoor and outdoor pollution, occupational dust and fumes, poorly managed airflow obstruction, and smoking~\citep{Yang}. In urban environments, pollution is often characterised by low average levels interrupted by short-term spikes, which are linked to asthma and COPD exacerbations requiring emergency care~\citep{Woodward}. Extreme atmospheric events---such as heatwaves, dust storms, and floods---also worsen asthma outcomes, especially in children and women~\citep{Makrufardi}. As climate change continues to increase the frequency of these events and overall pollution levels, the associated health risks are expected to grow significantly~\citep{PinhoGomes,Tran2022,Tran2023}. The respiratory system is especially vulnerable to air pollution, serving as the first point of contact for inhaled noxious agents and allergens. Common pollutants---including particulate matter (PM), nitrogen oxides (\ch{NOx}), sulphur dioxide (\ch{SO2}), ozone (\ch{O3}), and volatile organic compounds---can impair lung function and trigger inflammation in the airways, mediated by both neutrophilic and eosinophilic pathways~\citep{Montgomery}. While these effects are more pronounced in individuals with asthma, studies have shown that even healthy individuals exhibit signs of airway inflammation following exposure to these pollutants~\citep{McCreanor,Shi}. 

PM is categorised by aerodynamic diameter in three size fractions: coarse (PM$_{10}$, $\leqslant$ \SI{10}{\micro\meter}), fine (PM$_{2.5}$, $\leqslant$ \SI{2.5}{\micro\meter}), and ultrafine (PM$_{0.1}$, $\leqslant$ \SI{0.1}{\micro\meter}). Fine and ultrafine particles pose greater health risks due to their ability to penetrate deeply into the small airways and alveolar regions of the lung. Once deposited, these particles can cross the alveolar-capillary barrier, enter the bloodstream, and exert systemic effects on distant organs~\citep{Chen2022}.  Climate change may alter the physicochemical properties of particulate matter, influencing its size distribution and deposition patterns in the respiratory tract, thereby further increasing the risk of developing respiratory diseases~\citep{Tran2023,Kido}. 

Current strategies to mitigate air pollution exposure are broad and non-specific, offering limited protection for individuals who may be especially vulnerable~\citep{Agache, Carlsten}. Crucially, there are no tools that enable people to assess their personal risk in real-time environmental contexts~\citep{Becker}. Advances in wearable sensing and artificial intelligence offer an opportunity to close this gap. Personal sensing technologies now allow continuous collection of physiological and behavioural data, with demonstrated benefits across a range of health domains---including early detection of seizures, mood disorders, neurodegenerative conditions, and pulmonary diseases~\citep{Wang,SilveraTawil,Sameh,Whitlock}. In asthma care, mobile health (mHealth) systems such as Brisa~\citep{Cook}, the MASK-air app~\citep{Sousa-Pinto} and the MyAirCoach app~\citep{Khusial} are beginning to demonstrate their potential. Brisa, a chatbot developed to support asthma self-assessment and self-management, was perceived as helpful by patients; however, users expressed a need for greater conversational depth and improved personalisation~\citep{Cook}. The MASK-air app provided insights into patients' real-world medication use, giving physicians the opportunity to emphasise the importance of the treatment plan and how it maximises effectivity of medication~\citep{Sousa-Pinto,Avragim}. The MyAirCoach app, which monitors indoor air quality and physical activity, has been associated with improved symptom control and enhanced quality of life in patients with asthma~\citep{Khusial}. Wearables have also been shown to increase physical activity in individuals with COPD, though results for exacerbation detection remain mixed~\citep{Shah}. Integration with environmental data streams---as in the PulsAir app---further enhances user engagement and provides context-aware insights~\citep{Ottaviano}. Combining health data from wearables with environmental data sources (e.g., air quality, meteorology, and traffic) enables a more complete view of an individual's exposure profile~\citep{Khusial, Ottaviano,Helbig,Dang}. Concurrent breakthroughs in AI---particularly in deep and transfer learning---have further improved our ability to extract meaningful patterns from such heterogeneous, high-dimensional data~\citep{LeCun,Guo}. For example, neural networks trained on wearable data have been used to model personalised cardio-respiratory dynamics during exercise~\citep{Nazaret}. However, most sensing applications have focused on diabetes, cardiovascular health, and general wellness~\citep{SilveraTawil}, and their use in air pollution risk prediction remains limited.

We hypothesise that individual respiratory responses to air pollution---such as changes in breathing and heart rate---can be predicted using AI models trained on personal historical exposure data and physiological signals from wearables. This data-driven approach addresses the limitations of traditional population-level mitigation strategies and opens the door to anticipatory, personalised environmental health monitoring.

Building on our prior work in the INHALE project~\citep{INHALE}, which assessed pollution impacts across biological scales from the cell to the urban environment, AI-Respire~\citep{AI-Respire} advances personalised prediction of pollution-induced physiological responses. We present a modular, cloud-based workflow that integrates: (i)~physiological and pollution data from the INHALE cohort; (ii)~real-time smartwatch-derived health data via the BreathBot app; (iii)~location- and time-matched environmental data from OpenWeather~\citep{OpenWeather}; and (iv)~cross-sectional clinical data from the U-BIOPRED cohort~\citep{Shaw}.

\begin{figure}[htbp]
\centering
\includegraphics[width=0.9\textwidth]{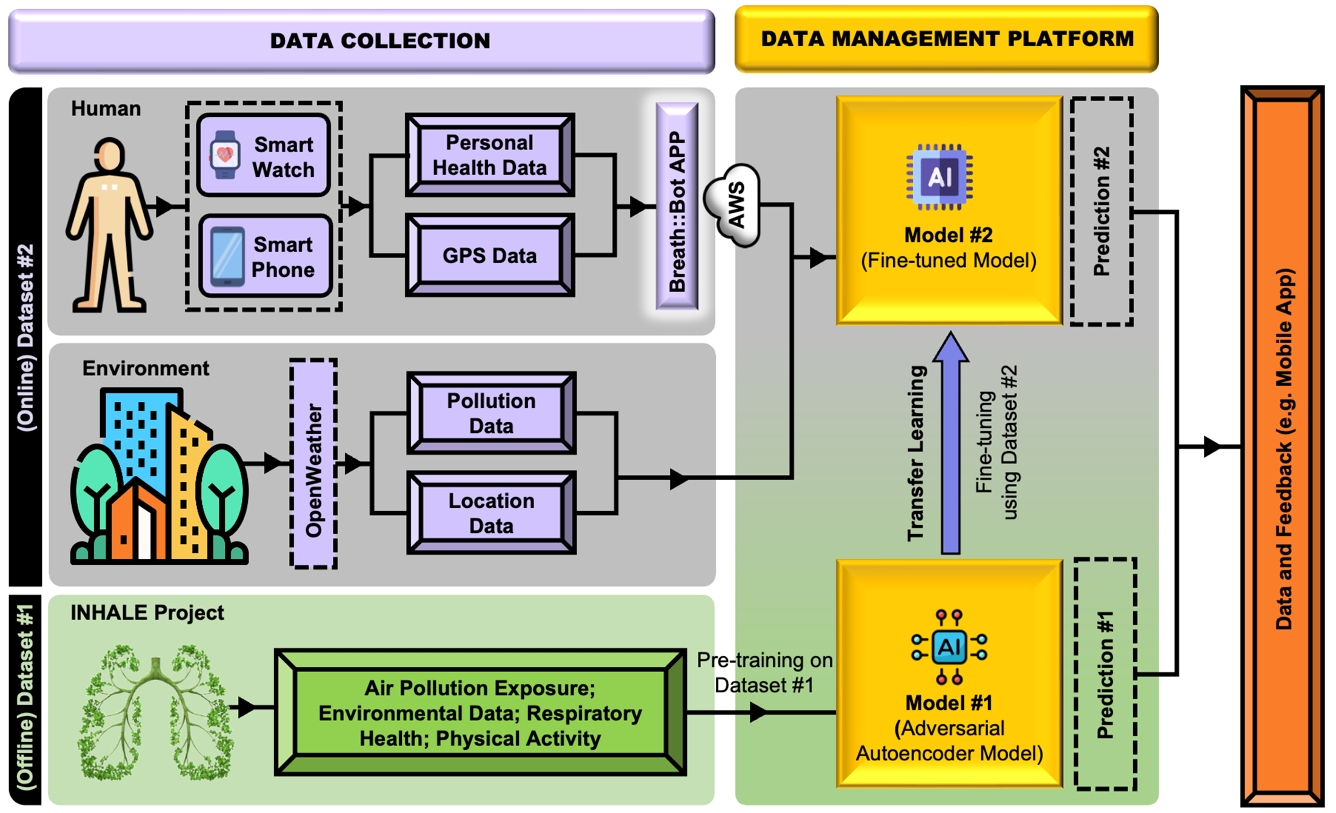}
\caption{\label{fig:workflow}Overview of the AI-Respire data pipeline and model development workflow. This schematic illustrates the end-to-end architecture developed in AI-Respire for predicting personalised health responses to air pollution. The pipeline integrates two datasets: (1)~an offline dataset from the INHALE project containing air pollution exposure, respiratory health, environmental, and physical activity data; and (2)~an online dataset collected in real-time via the BreathBot mobile app, including smartwatch-derived physiological data, GPS data, and pollution data synchronised through OpenWeather. Model \#1, an Adversarial Autoencoder (AAE), is pre-trained on the INHALE dataset to learn general patterns of pollution-health interactions. Using transfer learning, Model \#1 is then fine-tuned with personalised data (Dataset \#2) to yield Model \#2, which captures individual-level responses. Both models are deployed within a secure Data Management Platform (DMP), supporting predictions that feed back to the user (e.g., via the mobile app) for potential health insights or interventions. The system architecture ensures continuous data flow via secure AWS channels and supports scalable, ethical AI-driven personalised health monitoring.}
\end{figure}

The core component is an Adversarial Autoencoder (AAE)~\citep{Makhzani}, initially trained on INHALE data and then fine-tuned using smartwatch data through transfer learning to adapt to individual-specific patterns. To evaluate the clinical relevance of AI-predicted physiological shifts, we analysed U-BIOPRED data to determine whether model-simulated elevations in heart and breathing rate were associated with asthma burden~\citep{Zein}  or early signs of airway dysfunction. This article presents an end-to-end, AI-powered framework for personalised health response prediction to air pollution, detailing the underlying model architecture, multi-source data integration, and simulation-based validation---including cross-cohort analyses to assess the clinical relevance of model-predicted physiological shifts. Figure~\ref{fig:workflow} illustrates the full pipeline.

The paper is organised as follows: the next section 
describes the datasets, detailing the sources and structure of physiological and environmental datasets. The methodology section introduces the Data Management Platform (DMP) and Analytical Environment (AE) used to support secure and scalable data integration; the methods for data processing and feature engineering; and the development of AI models using both INHALE and smartwatch datasets. The results section evaluates the model’s performance, including simulation experiments to assess physiological responses to varying pollution levels, and outlines the application of transfer learning for personal health data. Finally, the paper concludes with a discussion of workflow implementation, current limitations, and future directions.

\section{Datasets}\label{sec:datasets}

\subsection{\textit{BreathBot:} Personal Health}
BreathBot is a custom mobile application (app) developed in the AI-Respire project~\citep{AI-Respire} to collect real-time health, location, and environmental data. It integrates with wearable devices and apps (e.g., Garmin, Fitbit, Apple Health, Android Health Connect) to retrieve physiological and self-reported metrics such as heart rate, respiratory rate, and, where supported, blood pressure, electrocardiogram and weight. Data acquisition occurs via Application Programming Interfaces (APIs) and scheduled webhooks, with continuous timestamped tracking. The app also records geolocation via a GPS plugin, updating when the Data Collection user moves beyond \SI{200}{m}, enabling spatial health analyses. Weather and air quality data, synchronised hourly with OpenWeather, are matched to user location, providing the environmental context for each health measurement. All data are securely transmitted to Amazon Web Services (AWS) and stored in encrypted databases. Regular secure transfers to a central DMP enable integrated, scalable, and GDPR-compliant analysis. Security protocols include multi-factor authentication, role-based access control, and transport layer encryption. Each API call is token-authenticated to preserve data integrity across the system lifecycle. 

\subsection{Environmental Data Collection}
Environmental data, including air quality and weather variables, were obtained from OpenWeather~\citep{OpenWeather}, a global provider offering high-resolution real-time and historical datasets. Parameters such as temperature, humidity, wind speed, and pollutant concentrations (PM$_{0.1}$, PM$_{2.5}$, PM$_{10}$, \ch{SO2}, \ch{NO2}, \ch{O3} and \ch{CO}) were accessed via the One Call~3.0 and Air Quality APIs. Data were linked to physiological readings using GPS coordinates and timestamps from the BreathBot and INHALE datasets.

\subsection{Overview of INHALE Dataset}
The INHALE~\citep{INHALE} dataset contains data from 59 participants aged~20 to~75 years. These participants include~33 non-asthmatic and 26~mild asthmatic individuals, each equipped with wearable sensors: the AIRSpeck~\citep{Arvind2016} and RESpeck~\citep{Arvind2019} devices. These capture information on air pollution exposure, respiratory health, and physical activity. Participants were monitored for two, two-week periods; one in summer and the other in winter. 

The AIRSpeck measures PM concentrations and ambient conditions at 30-second resolution via: (i)~an optical particle counter which measures particle counts across 16 size bins (grouping them into PM$_{0.1}$, PM$_{2.5}$ and PM$_{10}$ size fractions), and (ii)~temperature and relative humidity sensors. The RESpeck is a tri-axial accelerometer that is attached to each participant's chest, and is validated in measuring respiratory rate, also measuring physical activity level, step count and classifying movement (e.g., sitting, standing, lying down), at one-minute intervals. Additional weather and pollution data were supplemented from OpenWeather for corresponding timestamps and GPS location to enrich the dataset. 

Participants resided mainly in West London, with some located outside the area. The dataset focuses on the degree of exposure to environmental pollution and its potential respiratory effects. For this pilot study, we used data from 10 non-asthmatic INHALE participants. Hourly data was available, for each subject, two weeks of readings were available across both summer and winter periods; enabling the analysis of seasonal variations and environmental impacts on respiratory health. Table~\ref{INHALE-data} summarises the primary features included in the analysis.\\

\begin{table*}[htbp]
\footnotesize\sf\centering
\caption{Summary of INHALE dataset features used for model training. Hourly aggregates of physiological and environmental data were obtained from 10 non-asthmatic participants over two-week periods in winter and non-winter months. The table lists the description of each feature, units, value range, data type and source (RESpeck, AIRSpeck, or OpenWeather), supporting analysis of pollution exposure and respiratory health.\label{INHALE-data}}
\rowcolors{2}{gray!10}{white}
\begin{tabular}{p{2cm}p{3cm}p{2cm}p{2cm}p{2cm}p{3.2cm}}
\toprule
\textbf{Feature} & \textbf{Description} & \textbf{Units} & \textbf{Data type} & \textbf{Range} & \textbf{Source} \\
\midrule
Timestamp       & Date and time of the measurement with hourly resolution	& --- & datetime & --- & AIRSpeck/RESpeck\\
Breathing Rate  & Average respiratory rate of the subject per minute & bpm (breaths/min)	& float	& 8 -- 30	& RESpeck\\
Breathing Rate STD & Standard deviation of the respiratory rate per minute & bpm (breaths/min) & float & 0 -- 5 & RESpeck\\
Activity Level	& Intensity of the subject's activity &--- & float & 0 -- 10 & RESpeck \\
Step Count      & Number of steps taken by the subject per minute & steps & integer	& 0 -- 120 & RESpeck\\
Temperature     & Ambient temperature & \si{\celsius} &	float & -5 -- 40 & AIRSpeck/OpenWeather\\
Humidity        & Relative humidity & $\%$ & float & 0 -- 100 & AIRSpeck/OpenWeather\\
PM2.5 Concentration & Particulate matter $\leqslant$\SI{2.5}{\micro\meter} in diameter &	\si{\micro\gram\per\cubic\meter} & float & 0 -- 500 & AIRSpeck/OpenWeather\\
\bottomrule
\end{tabular}
\rowcolors{0}{}{}
\end{table*}

\subsection{Overview of U-BIOPRED Dataset}
We analysed data from the U-BIOPRED adult cohort~\citep{Shaw}, a well-characterised dataset comprising healthy volunteers (HV), mild-to-moderate asthma (MMA), and severe asthma (SA) patients. The severe asthma group was further stratified into smoking (SA.s) and non-smoking (SA.ns) subgroups. Demographic, clinical, and physiological parameters were available for each participant, including heart rate (HR) and respiratory rate (BR) measured at baseline. For individuals with asthma, we assessed whether elevated HR and BR were associated with asthma burden score, a validated composite index reflecting healthcare utilisation, symptom control, and quality of life impairment~\citep{Zein}. Among healthy volunteers, we explored associations with Fractional Exhaled Nitric Oxide (FeNO), a recognised biomarker of airway inflammation.

\section{Methodology}\label{sec:methodology}
\subsection{Data Management Platform and Analytical Environment}
All data were stored, integrated, and analysed within a secure Data Management Platform and Analytical Environment (DMP-AE) developed by the Data Science Institute at Imperial College London. This cloud-based infrastructure supports standardised ingestion, harmonisation, and analysis of heterogeneous datasets---including physiological, environmental, and clinical data. The AE includes a high-performance compute environment with support for Python, R, MATLAB, and Jupyter Notebooks, enabling interactive model development, time-series processing, and visualisation. Model training (including that of the adversarial autoencoder and transfer learning) was performed using GPU-enabled compute nodes, allowing efficient handling of high-dimensional, multi-source data. All scripts and workflows were containerised to ensure reproducibility across development and deployment environments. Security protocols include multi-factor authentication, role-based access control, and encrypted data transfer and storage, ensuring GDPR compliance. Robust backup systems and automated logging ensure data integrity, resilience, and traceability throughout the research lifecycle.

\begin{figure}[htbp]
\centering
\includegraphics[width=0.45\textwidth]{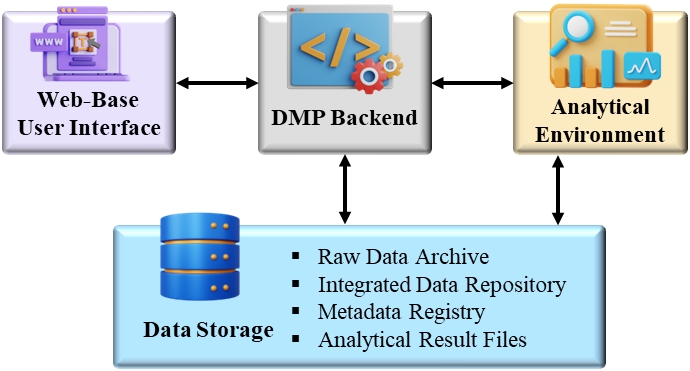}
\caption{\label{fig:DMP}Data Management Platform (DMP) architecture in the AI-Respire project. It includes four core components: a secure web interface, backend processing engine, analytical environment, and data storage. The storage system supports raw and integrated data, metadata, and model outputs, enabling scalable, traceable, and secure AI-driven health analysis.}
\end{figure}

The platform architecture, shown in Figure~\ref{fig:DMP}, demonstrates the key components and their interconnections, including: (i)~data storage, (ii)~backend operations, (iii)~the analytical environment, and (iv)~the web-based interface. This setup enables efficient data exchange, real-time synchronisation, and secure access to support health-related research. The infrastructure adheres to standards for data security, quality management, and accessibility. Hosted within an ISO-compliant data centre, the platform ensures information security by leveraging Imperial College London’s advanced AI Cloud infrastructure, which provides a substantial 2-petabyte (PB) hybrid object storage. This storage capacity supports large-scale, high-dimensional data required for our detailed pollution and health analysis.

\subsection{Data Integration and Model Development}
This study uses three main data streams: (1)~physiological and environmental data from the INHALE cohort, (2)~personal smartwatch data and (3)~publicly available pollution and weather data via the OpenWeather API. The INHALE data subset used for this analysis included 10 healthy participants monitored using AIRSpeck (air quality) and RESpeck (respiratory rate and activity) sensors across two seasonal periods: Summer (P1) and Winter (P2) (Table~\ref{INHALE-data}). Data were recorded hourly and geotagged with GPS coordinates, enabling alignment with external datasets. This dataset was temporally and spatially matched with OpenWeather pollution and meteorological records, retrieved using timestamps and GPS coordinates. Personal data were collected independently from a single user over 8~months via Apple Health (heart rate, step count) and Google Maps Timeline (location). This dataset was temporally and spatially matched with OpenWeather pollution and meteorological records, retrieved using timestamps and GPS coordinates. All datasets were integrated into the secure, GDPR-compliant DMP-AE, supporting harmonised storage and scalable analysis.

\subsection{Data Cleaning}
Initial exploratory analysis revealed temperature outliers in the INHALE dataset, with approximately 50\% of participant records containing implausible values (e.g., \SI{40}{\celsius} in January). Possible explanations are that the protruding temperature sensor became trapped close to the particpants' bodies or that the additional heat came from the device itself, which had no internal fan. This issue higlights the complexity of designing wearable sensors to ensure that they measure what what is desired. The outliers were corrected by capping values to realistic regional bounds informed by typical historical weather patterns in London. The bounds were in the range \SI{-5}{\celsius} to \SI{35}{\celsius}. For about 5\% of participants, breathing rate values below 8 breaths per minute were observed. These were flagged as physiologically implausible and replaced with the median value for the respective participant and seasonal period. 

OpenWeather's pollution data was available hourly, whereas their historical weather data was available every minute. In order to retain as much fine-grained information as possible, the pollution data was interpolated to give observations every minute. The interpolation was performed within each participant-period group (P1: Summer, P2: Winter). Any remaining missing values were filled using the group-specific mean. This approach preserved individual variability and seasonal trends while avoiding distortion from global averages.

\subsection{Data Preprocessing}
\subsubsection{Normalisation}
Normalisation is a critical step in data preprocessing, particularly when working with time-series data that exhibit varying scales and ranges across features. To prepare the data for model training, Min-Max scaling was employed to normalise the data, ensuring that all features fall within a consistent range, typically between 0 and 1. This normalisation helps to ensure that no single feature disproportionately influences the model due to its scale, which could otherwise slow convergence or lead to unstable training. In neural networks, where input features should be within the same scale, this normalisation allows for efficient gradient descent and faster convergence. The Min-Max scaler was fitted exclusively on the training data to prevent data leakage, and the same transformation was then applied to the validation and test sets, maintaining consistency across all datasets. This approach ensures that the model is evaluated in a fair and unbiased manner. After normalisation, the scaled features were re-integrated with their corresponding identifiers (e.g. patient ID, timestamp, period) for subsequent analysis. Categorical variables (e.g., time of day, day of week) were encoded cyclically using sine and cosine transformations to preserve their temporal structure.
Min-Max scaling rather than alternative scaling methods such as Z-score normalisation was chosen for its ability to bring all features within a bounded range, which is particularly beneficial when using neural networks with activation functions such as sigmoid or tanh~\citep{Dubey}. Moreover, the potential influence of outliers was addressed during the exploratory data analysis phase, ensuring that the scaling process remains robust.
\subsubsection{Time Series Construction}
To model sequential patterns in the data, time-series windows were constructed using a sliding window approach. The data was reshaped into a three-dimensional tensor with dimensions: \texttt{[samples, ntimes, features]}. After empirical testing and hyperparameter optimisation, a window length of 8-time steps (\texttt{ntimes=8}) and an overlap of 7 time steps between adjacent windows was selected. Given a dataset of 100 timesteps and 29 features, the resulting 3D tensor would have dimensions \texttt{[93, 8, 29]}. This construction ensures the sequence of events and timing between them are preserved, enabling the model to learn patterns across time. Small overlaps retain more temporal information, which can enhance model performance for datasets with rapid changes but may increase computational cost. Conversely, larger overlaps prioritise efficiency but risk omitting critical information. For the INHALE dataset, where pollution and health metrics exhibit temporal variability, this method is well-suited to capture both short-term fluctuations and long-term trends.  

\subsection{Feature Engineering}
The INHALE dataset includes features of air pollution exposure from size fraction particle counts. Cyclical features such as hour of day and day of week were encoded using sine and cosine transforms to reflect their periodic nature. This technique allows the model to capture recurring patterns, such as diurnal variations in pollution levels or weekly behavioural cycles, that might influence health responses. These transformations were applied to all time-dependent features, ensuring that periodic trends were represented in a format compatible with the model architecture. The effectiveness of these engineered features will be evaluated in subsequent stages of model training and validation. Additionally, a synthetic heart rate feature was derived using the approximation and breathing rate: HR $\approx$ 4 $\times$ BR, where 4 is used as an approximate value for the pulse-respiration quotient~\citep{Kalauzi} for healthy adults. This enabled us to simulate real-world wearable data, where heart rate is commonly available even when respiratory data is not. To simulate wearable-derived data variability, we applied $\pm$10\% noise to the derived HR values, consistent with real-world measurement variance. Finally, features with low signal or redundancy (e.g., air particle bin counts) were excluded to reduce dimensional noise. 

The addition of heart rate was motivated by its potential to strengthen the capacity of the model to capture physiological responses to pollution exposure. Although, the use of a fixed pulse-respiration quotient introduces assumptions that may not fully account for individual differences due to factors such as age, fitness level, or underlying health conditions. Future iterations could refine this feature by incorporating actual observed heart rate data or subject-specific adjustments where available.

\subsection{Model Architecture}
\begin{figure*}[htbp]
\centering
\includegraphics[width=0.85\textwidth]{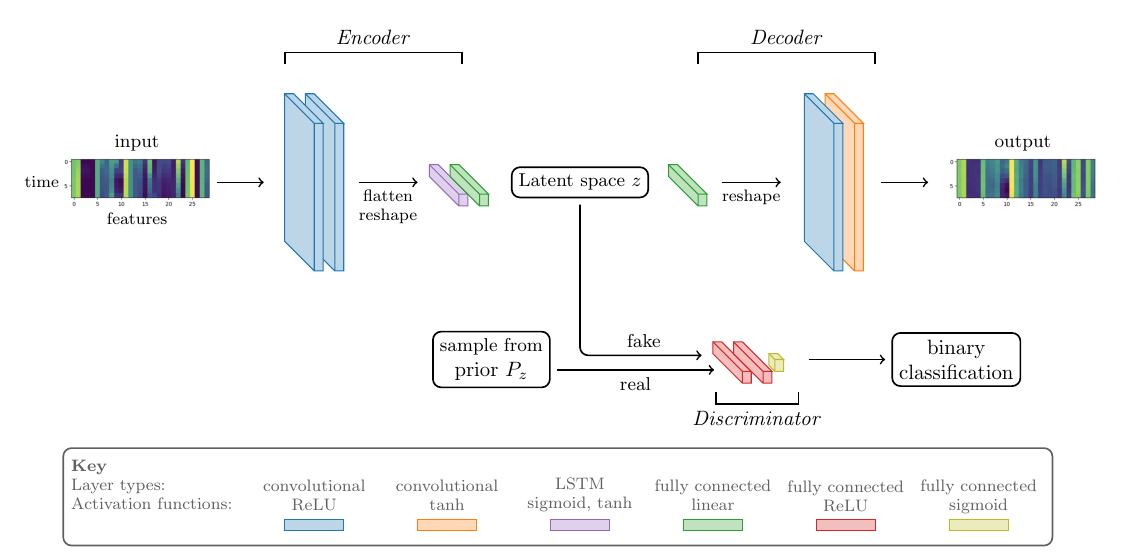} 
\caption{\label{fig:AAE}Overview of the AAE architecture for time-series prediction of pollution-linked health outcomes. The adversarial training is achieved through a discriminator which takes input either by sampling a prior distribution or from the latent space. Colour indicates the type of layers, the key for which is seen at the bottom of the image.} 
\end{figure*}

The AAE framework forms the foundation of the proposed model, integrating the principles of autoencoders with adversarial learning to achieve robust feature representation and well-structured latent spaces. To enhance its predictive capabilities, the architecture is extended with Long Short-Term Memory (LSTM) layers to capture temporal dependencies inherent in the time-series data~\citep{Olah}. Moreover, convolutional layers are incorporated to model spatial correlations effectively~\citep{Aghdam}. Given the limited size of the training dataset, attention mechanisms were excluded to avoid overfitting and maintain computational efficiency. This hybrid architecture enables the model to capture both short-term dynamics and long-range spatial patterns.
Figure~\ref{fig:AAE} provides a schematic representation of the model architecture, highlighting its key structural components: an encoder, a latent space regularisation mechanism, a decoder, LSTM layers to capture temporal dependencies and convolutional layers for spatial correlations. Complementing these components are the objective or loss function, the training process and the model outputs. The input consists of 8 time levels (i.e. an 8-timestep window), where each time level contains 29 features. It is reshaped to a 3D tensor \texttt{[samples, 8, 29]}. The encoder includes two 2D convolutional layers with ReLU activations, followed by an LSTM layer with sigmoid and tanh activations, and a fully connected layer with a linear activation function projecting to a 1000-dimensional latent space ($z$). The decoder mirrors this process to some extent, with a fully connected layer followed by two transposed convolutional layers using ReLU and tanh activations to reconstruct the output. A discriminator network, composed of three fully connected layers, the first two with the ReLU activation function and the final layer with a sigmoid function. The discriminator enforces adversarial regularisation by distinguishing real samples from the prior $P_z$. During training (20 epochs, Adam optimizer), the model learns both reconstruction and adversarial losses. This architecture supports both prediction and inpainting of missing data across time by iteratively passing sequences through the trained model. Colour-coded blocks indicate the layer types and activation functions, see Figure~\ref{fig:AAE}. 

The encoder maps the input data  $x$ to a latent representation $z$ in a lower-dimensional space while preserving critical features. This process is defined as:
\begin{equation}
z=f_{\text{enc}}(x;\theta_{\text{enc}})\,.
\end{equation}
To impose structure on the latent space, a discriminator $D$ adversarially enforces similarity to a prior distribution $P_z$ (e.g. a Gaussian distribution). The adversarial loss is given by:
\begin{equation}
\mathcal{L}_{\text{adv}} = - \E_{z\sim q(z)} [ \log D (z) ] - \E_{z\sim P_z} [ \log \left(1 - D (z) \right) ] 
\end{equation}
where $q(z)$ is the aggregated posterior distribution of the latent space. The decoder attempts to reconstruct the input $\hat{x}$ from the latent representation $z$,
\begin{equation}
\hat{x} = G(z) = f_{\text{dec}} (z;\theta_{\text{dec}})\,,
\end{equation}
by optimising a reconstruction loss which is defined as:
\begin{equation}
\mathcal{L}_{\text{rec}} = \E_{x\sim P_{\text{data}}} \|x - \hat{x} \|^2\,.
\end{equation}
The total loss function combines the reconstruction loss, adversarial loss and any additional regularisation terms (not shown):
\begin{equation}
\mathcal{L}_{\text{total}} = \mathcal{L}_{\text{rec}} + \lambda_{\text{adv}} \mathcal{L}_{\text{adv}}
\end{equation}
where $ \lambda_{\text{adv}}>0$ balances the contribution of the adversarial loss with the reconstruction loss. We set $ \lambda_{\text{adv}}=1$, which we found to provide a good balance between reconstruction accuracy and latent space regularisation. 

To model temporal patterns, the latent representation $z_t$ at each timestep $t$ is processed using LSTM layers. The LSTM captures dependencies across time windows, linking $z_t$ with earlier steps through its hidden states $h_t$ and cell states $c_t$:
\begin{align}
i_t &=\sigma ( W_i z_t + U_i h_{t - 1} + b_i )\\
o_t &= \sigma ( W_o z_t + U_o h_{t - 1} + b_o )\\
c_t &= f_t \odot c_{t-1} + i_t \odot \tanh (W_c z_t + U_c h_{t-1} + b_c )\ \\
h_t &= o_t \odot \tanh (c_t)
\end{align}
where $\sigma$ is the sigmoid activation function, $\odot$ denotes element-wise multiplication represented by the Hadamard product, and $W$, $U$, and $b$ are learnable weights and biases with appropriate subscripts for each gate or component. As shown in Figure~\ref{fig:AAE}, these activations are specific to the LSTM module (purple), while the decoder uses ReLU activations (red) for image reconstruction. The use of standard LSTM gating ensures temporal continuity in the latent representation, while ReLU facilitates non-linear decoding of the output signal. See Crilly~et al.~\citep{Crilly} for general information on AI and neural networks, and Wassan et~al.~\citep{Wassan} for more information on aspects of training neural networks and activation functions. To improve the stability of the predictions, a technique known as unrolled or rollout training was used~\citep{Um2020,Kochkov2021}. During training, instead of predicting one future time level and calculating the loss function on the difference of this with the ground truth, predictions are made for a sequence of time levels and the loss is based on the difference between the ground truth and predictions for the entire sequence.

Data inpainting is a technique used to reconstruct missing or corrupted portions of data by leveraging contextual information from the surrounding regions~\citep{Xie}. In this study, inpainting is used to make predictions taking the values at future time levels to be the missing values~\citep{HeaneyCE,GuoD}. Consider an AAE that has been trained to accept a sequence of~8 time levels $\{x_1, x_2,\ldots, x_8\}$. The output of the AAE will consist of a new sequence of values,  $\{\hat{x}_1, \hat{x}_2,\ldots, \hat{x}_8\}$, that will be close to the original values if the AAE has been well trained. To use an AAE for prediction, seven known values can be used $\{x_1, x_2,\ldots, x_7\}$ to predict the next value in the sequence, $x_8$, by (1)~making an initial guess for $x_8$ (using noise or by assigning it the value of $x_7$); (2) passing $\{x_1, x_2,\ldots, x_8\}$ into the AAE; (3)~obtaining an improved estimate for ${x}_8$ from the output of the AAE $\hat{x}_8$. Steps~(2) and~(3) represent inpainting. The value for $x_8$ is updated and can be passed back into the AAE with the original seven time levels. After a certain number of iterations or once the value for $x_8$ has converged, we move on to the next time level. An initial estimate for $x_9$ and the seven previous values are fed into the AAE, $\{x_2, x_3,\ldots, x_9\}$ and the output of the AAE yields an improved estimate, $\hat{x}_9$. The iteration continues until convergence and the time-stepping process continues until the desired number of time levels has been reached. In this example, there is one variable in the input ($x$) whereas in the models reported in this study, three variables are used, pollution, weather and heart rate.

\section{Results}\label{sec:results}

\subsection{Model Training and Evaluation}
The training and prediction workflow for the proposed AAE model involves three main stages: training, inpainting-based prediction, and evaluation. During training, the encoder and decoder learn feature representations and reconstruct the input data, while the discriminator ensures latent space regularisation using adversarial loss. The model is optimised iteratively using a combination of reconstruction and adversarial objectives. For prediction, future values are iteratively reconstructed using inpainting, where the model leverages its learned representations to infer plausible data points. Finally, evaluation metrics such as MSE and visualisation techniques are used to assess the performance of the model. A detailed outline of the process is provided in Listing~\ref{fig:algorithm}. 

\begin{figure*}
\centering
\begin{lstlisting}[caption={Training and prediction of the AAE with performance evaluation}, label={fig:algorithm}, language=python]
# Input: Training data(DTrain),Test data(DTest)
# Output: Predicted values for missing data, Evaluation metrics

Initialise model parameters (thetaEnc, thetaEnc, thetaDis)
Set hyperparameters: maximum iterartions (NEpochs), noise level (epsilon), inpainting iterations (NInpaint)
Load training data into (train_loader), test data into (test_loader)

--- Training Phase ---
for each epoch (e=1, NEpochs) do
  for each batch (xBatch, yBatch) in (train_loader) do  
    # Encoder-Decoder forward pass
    z = Encoder(xBatch; thetaEnc)
    xRecon = Decoder(z; thetaDec)
    Compute reconstruction loss (Lrecon = MSE(xBatch, xRecon))

    # Latent space regularisation
    Generate zReal by sampling the prior latent distribution 
    Generate zFake using the encoder
    Compute adversarial loss (Ladv) using (Discriminator(zFake, zReal; thetaDis))

    # Update parameters
    Update (thetaEnc, thetaDec) using (Ltotal = Lrecon + lambda_adv Ladv)
    Update (thetaDis) using (Ladv)
  end for
end for

--- Prediction Phase with Inpainting ---
for timestep (t=1, T) do
  xSample = test_loader[t]          # Get 7 previous time levels
  xPred = xSample[7] + epsilon      # Approximate future time level with the previous value
                                    
  # Iterative inpainting
  for iteration (i=1, NInpaint) do
    xInput = [xSample, xPred]       # Append current best guess for future time to previous time levels                                        
    xOutput = Decoder(Encoder(xInput; thetaEnc); thetaDec))
    xPred = xOutput[8]              # Updated prediction for the future time level 
  end for
  Store final prediction: (xFinal[t] = xPred)
end for

--- Evaluation ---
# Compute evaluation metrics (Mean Squared Error, R2) between (xFinal) and ground truth (yTruth)
Mean Square Error = MSE(xFinal, yTruth)
MSE_res = MSE(xFinal, yTruth)
MSE_tot = MSE(MEAN(xFinal), yTruth)
R2 = 1 - MSE_res / MSE_tot

\end{lstlisting}
\end{figure*}

\begin{figure*}[htbp]
\centering
\includegraphics[width=0.7\textwidth]{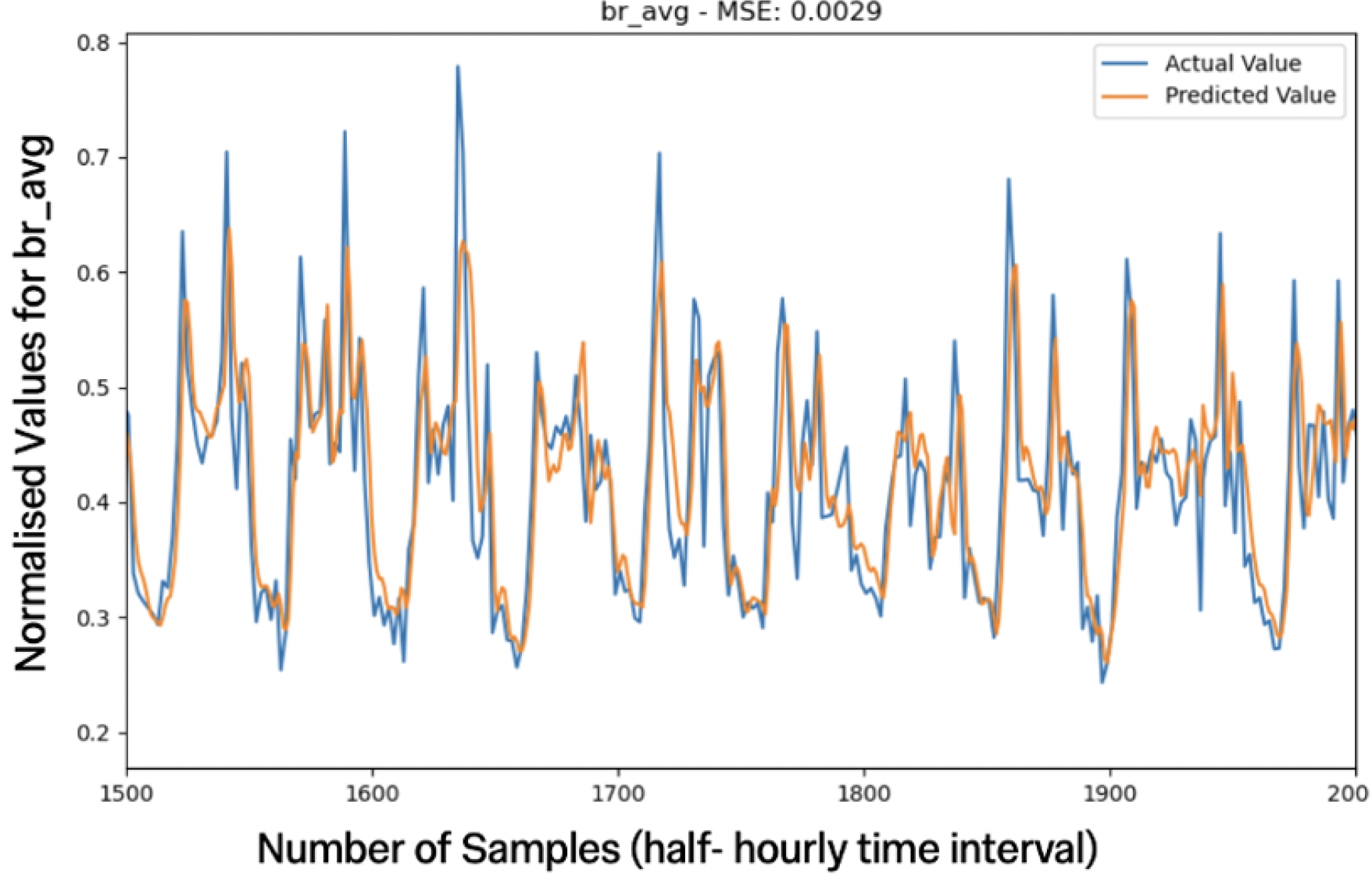}
\caption{\label{fig:results}Comparison of actual and predicted normalised breathing rate (\textsf{\small br\textunderscore{}avg}) values over half-hourly intervals.} 
\end{figure*}

Using real-world data obtained from a smartwatch, we investigated the ability of the model to reconstruct and predict normalised values of breathing rate (\textsf{\small br\textunderscore{}avg}). To generate the prediction, pollution, weather and heart rate data were taken from the test dataset for seven time levels. The value of these variables was calculated at the eighth time level using inpainting as described in the previous section. The time-stepping procedure continues until the desired number of predictions have been made. Figure~\ref{fig:results} shows the comparison between the predicted values and the actual values taken from the test dataset over a sample range, normalised for clarity. The results demonstrate that the predicted values (orange line) closely align with the actual values (blue line), capturing both short-term fluctuations and long-term trends. The performance of the proposed AAE was evaluated on a held-out validation set, where it achieved a final mean squared error (MSE) of 0.0029 in reconstructing normalised breathing rate values. During training, the reconstruction loss (MSE) decreased by more than one order of magnitude, from approximately 0.022 to below 0.001 over 40 epochs. This substantial reduction highlights the model’s ability to learn meaningful temporal structure in the input features. The final predictions closely aligned with actual values, capturing both short-term fluctuations and longer-term respiratory patterns---despite the absence of an explicit attention mechanism---demonstrating the effectiveness of the AAE architecture in modelling physiological time series.

\subsection{Effect of Pollution Levels on Respiratory and Cardiovascular Health}
\begin{figure*}[htbp]
\centering
\includegraphics[width=0.7\textwidth]{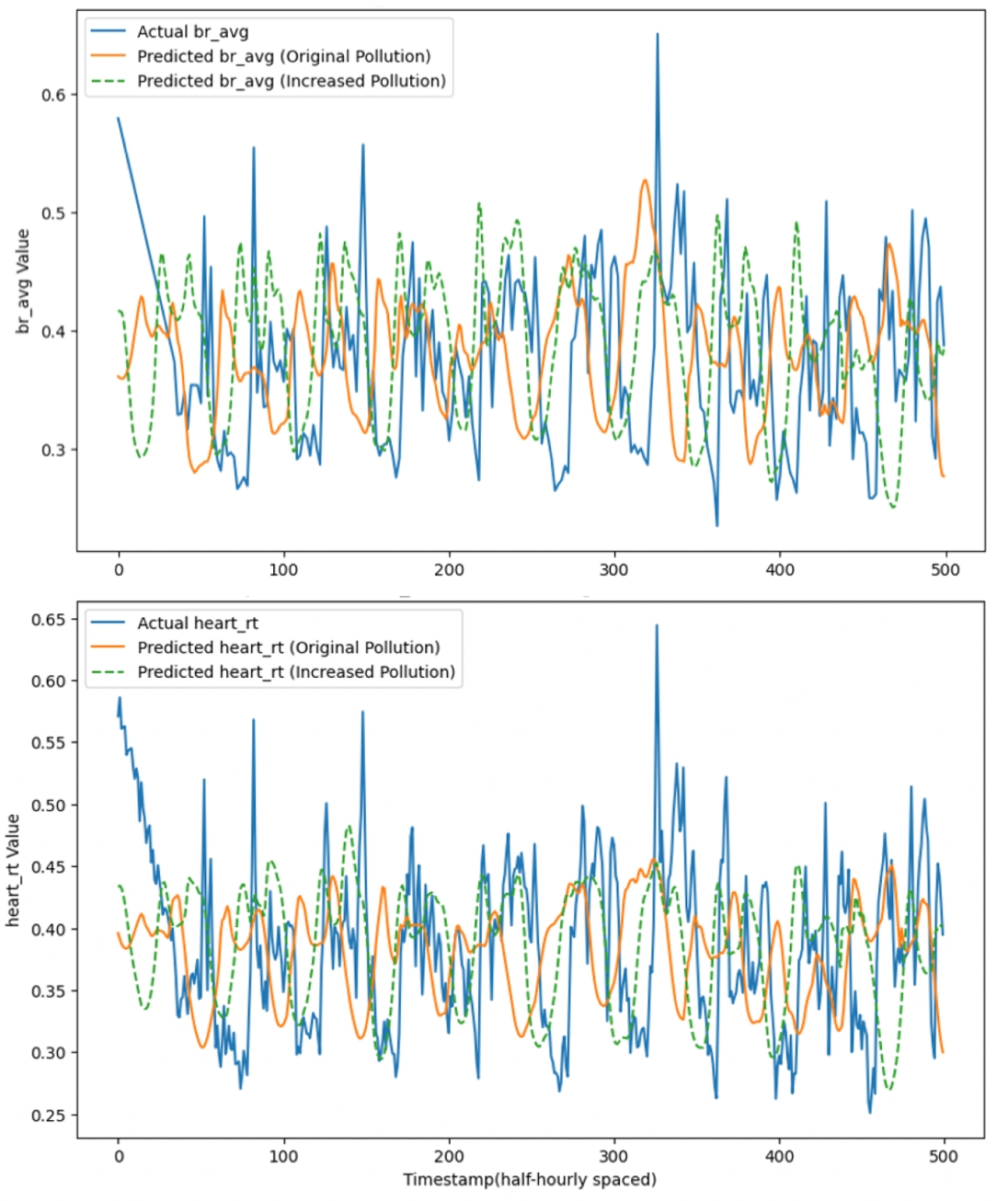}
\caption{\label{fig:more-results}Comparison of predicted and actual trends for average breath rate (top) and heart rate (bottom) under baseline and 100\% increased pollution levels.} 
\end{figure*}

The relationship between air pollution and physiological responses---specifically respiratory and cardiovascular metrics---was assessed using the proposed AAE model. This relationship is key, as, although the effects of air pollution are more pronounced in individuals with asthma, healthy individuals can exhibit signs of airway inflammation following exposure. Initial correlations between pollution levels and both average breathing rate (\textsf{\small br\textunderscore{}avg}) and heart rate (\textsf{\small heart\textunderscore{}rt}) were weak. To explore potential nonlinear effects, we simulated chronic pollution increases of 20\%, 50\%, and 100\% across key pollutants (PM$_{2.5}$, PM$_{10}$, \ch{NO2}, \ch{O3}, and \ch{CO}). Although a 100\% increase may appear extreme, such levels are reached episodically during wildfires, traffic surges, or weather inversions and are projected to occur on 10--20\% of days annually in urban areas due to climate change. Our simulation, which applied sustained increases across the complete time window, revealed that only the 100\% condition led to measurable physiological changes: breathing rate rose by $\sim$3.5\% and heart rate by $\sim$2.5\%, see Figure~\ref{fig:more-results}. While modest, these shifts may reflect early physiological stress in vulnerable individuals, reinforcing the importance of personalised monitoring and anticipatory risk modelling.

To contextualise the AI-predicted physiological shifts---namely, a 2.5\% increase in heart rate and a 3.5\% increase in breathing rate under high pollution scenarios---we conducted a cross-sectional analysis using the U-BIOPRED cohort. Rather than assessing within-person changes over time, we examined whether individuals who already exhibited these subtle elevations at rest differed clinically from their peers. This approach aimed to evaluate the potential health significance of small physiological deviations predicted by the model, even in the absence of acute exposure data.
Among people with asthma, those classified with elevated HR had significantly higher asthma burden score, even after adjusting for age, sex, and BMI. This implies that even modest increases in HR---well within normal physiological ranges---may be early indicators of disease burden. 

In healthy volunteers, elevated HR (defined similarly) was associated with higher levels of FeNO, a recognised biomarker of allergic airway inflammation, as well as subtle signs of irregular airflow. This aligns with prior panel studies, such as Shi et al.~\citep{Shi} who demonstrated that even short-term exposure to ambient PM$_{2.5}$ in healthy young adults significantly increased FeNO.

\subsection{Transfer Learning for Personal Health Data using Smartwatch Measurements}
To enhance the adaptability of the neural network initially pre-trained on the INHALE dataset, we applied transfer learning by fine tuning the model using personal health data collected over eight months from a smartwatch. The dataset comprises time-series data with 5-minute time intervals, including physiological metrics such as heart rate and step count, coupled with environmental factors such as weather and air pollution levels. The weather and pollution data were sourced from the OpenWeather API, providing a feature set for model training.

The transfer learning approach involved freezing the initial layers of the pre-trained model, preserving general feature representations, while fine-tuning the final two layers with the smartwatch dataset. This allowed the model to adapt to individual-specific patterns while leveraging previously learned pollution-health dynamics, improving efficiency and predictive performance. Consider a pre-trained model $f_{\text{pre}}(x;\theta_{\text{pre}})$ parameterised by $\theta_{\text{pre}}$ over $L$~layers with input features~$x$. A new model $f_{\text{adapt}}(x;\theta_{\text{adapt}})$ also with $L$~ layers is sought, which has the same weights as $\theta_{\text{pre}}$ for all but the final two layers. The weights of the final two layers of $f_{\text{adapt}}$ are determined by the smartwatch data through the following minimisation process
\begin{equation}
\theta_{\text{adapt}} = \argmin_\theta \mathcal{L} \left(y,f_{\text{adapt}}(x;(\theta^*,\theta)\,\right)
\end{equation}
where $\theta^*$ represents the frozen weights (from the first $L-2$ layers of $\theta_{\text{pre}}$) and $\theta$ represents weights of the final two layers. The smartwatch dataset was divided into training and test subsets, with 80\% used for training the final two layers and 20\% retained for evaluation. During training, the Adam optimiser was employed, with a learning rate of \num{e-4}, to minimise the loss and update the model parameters of the last two layers. 

The model achieved a mean squared error (MSE) of \num{4.24e-5} on the unseen 20\% of smartwatch data, closely aligning with the reconstruction loss observed during training ($<$0.001), suggesting strong internal consistency and adaptability to real-world temporal variability. Qualitative evaluation confirmed that the model accurately captured temporal patterns in physiological signals, including those modulated by environmental exposures. The successful application of transfer learning highlights the flexibility of the proposed approach. The ability to generalise across datasets with distinct characteristics makes it a promising tool for broader applications, such as personalised health monitoring and environmental health assessments. Future work will aim to explore other transfer learning strategies, such as freezing intermediate layers or using domain adaptation techniques, to further enhance performance on diverse datasets.

\section{Conclusions and Future Work}\label{sec:conclusions}

This study presents a novel AI-driven framework for predicting personalised physiological responses to air pollution by integrating wearable-derived health data with real-time environmental exposures. At its core, the predictive model—a generative Adversarial Autoencoder enhanced with LSTM and convolutional layers—successfully reconstructed time-series health signals and captured responses to pollution. The model achieved consistent performance and demonstrated adaptability across datasets of differing resolution and context via transfer learning on smartwatch data. 

To support this modelling pipeline, we developed a secure, cloud-based DMP-AE, enabling standardised ingestion, harmonisation, and analysis of heterogeneous data sources, including the INHALE cohort, personal wearables, and environmental APIs. The INHALE dataset, with co-located, high-resolution measurements of indoor/outdoor air quality and physiological metrics, offered a unique foundation for training. Transfer learning on personal smartwatch data confirmed the model’s ability to generalise across datasets with differing resolution, structure, and context. The framework maintained performance (MSE $\approx$ 0.003) without retraining the entire network, illustrating its adaptability to real-world, user-generated data.
Both the INHALE and smartwatch datasets demonstrated the feasibility of AI-based modelling of pollution-health relationships. While standard correlation analyses showed weak associations between PM levels and cardiorespiratory signals in the healthy INHALE cohort, our AI-based model successfully reconstructed and predicted physiological responses. This suggests that machine learning methods can uncover latent or nonlinear pollution–health interactions, which may be invisible to traditional univariate analysis—particularly in small or resilient populations. 
Our simulation, which applied sustained increases across the full-time window, revealed that only the 100\% condition led to measurable physiological changes: AI predicted breathing rate rose by $\sim$3.5\% and heart rate by $\sim$2.5\%. While modest, these shifts may reflect early physiological stress in vulnerable individuals, reinforcing the importance of personalised monitoring and anticipatory risk modelling. 

To contextualise the AI-predicted physiological shifts---namely, a 2.5\% increase in heart rate and a 3.5\% increase in breathing rate under high pollution scenarios---we conducted a cross-sectional analysis using the U-BIOPRED cohort. Rather than assessing within-person changes over time, we examined whether individuals who already exhibited these subtle elevations at rest differed clinically from their peers. 

In U-BIOPRED, individuals with elevated heart rate showed higher asthma burden scores~\citep{Zein} (asthmatics) or elevated FeNO and airflow irregularities (healthy volunteers), supporting the physiological relevance of the AI-predicted shifts. These findings suggest that even in the absence of diagnosed disease, small vital sign elevations may signal underlying physiological stress or early airway changes and highlights that the AI is producing shifts that correspond to measurable and relevant health outcomes in real data.

Together, our findings demonstrate the feasibility and utility of generative AI models for predicting individualised physiological responses to air pollution. Simulated pollution increases produced modest but measurable shifts in heart and breathing rates, which aligned with clinical and biological markers in external cohorts—supporting their potential as early, actionable digital biomarkers. While longitudinal studies are needed to establish causality, this cross-sectional benchmark underscores the plausibility and clinical relevance of using subtle cardiorespiratory changes for early risk stratification in both asthmatics and healthy individuals. Personalisation via transfer learning on smartwatch data further improved model adaptability to real-world settings. Beyond prediction, the supporting cloud-based infrastructure enables secure, standardised integration of offline and real-time multimodal data, laying a foundation for future personalised environmental health monitoring and precision interventions.

While promising, the study is not without limitations. The relatively small sample size and limited demographic diversity constrain the generalisability of our findings. Future work should prioritise inclusion of children, older adults, and individuals with asthma or cardiovascular disease---populations known to be especially vulnerable to pollution---to yield more robust and actionable insights. Enhancing model training through time-series generative modelling offers a promising avenue to address data scarcity, enabling the creation of synthetic datasets that retain meaningful temporal structure. Additionally, distinguishing the impacts of indoor versus outdoor pollution using integrated environmental and smart home sensor data could refine individual exposure estimates and support dynamic mitigation strategies. Ultimately, this framework has the potential to improve the quality of life for high-risk groups by enabling real-time, AI-guided responses to environmental challenges.

Building on the findings of this study, we identify several key aspects that warrant further investigation to enhance our understanding of pollution exposure and its health implications at an individual level:
\begin{itemize}
\item Identifying the main constituents of pollution that individuals are exposed to and quantifying their daily exposure levels. 
\item Assessing the environmental factors that influence pollutant levels in an individual’s surrounding and their role in modulating exposure. 
\item Establishing measurable health indices that can serve as reliable surrogates for an individual’s physiological response to air pollution. 
\item Investigating potential biomarkers of risk that could provide early indicators of adverse health effects.
\item Developing methods to assess both environmental exposure and individual-level respiratory and cardiovascular responses. 
\item Exploring predictive models that link exposure patterns to long-term health outcomes at an individual scale.
\end{itemize}
Addressing such questions in future research will contribute to a more comprehensive framework for assessing the health impacts of air pollution and guiding personalised mitigation strategies.

\vspace{28pt}
\hrule
\vspace{14pt}
\vspace{7pt}
\noindent\textbf{\textsf{Funding sources}}\\[5pt]
\small This work was supported by the following UKRI grants: AI-Respire, ``AI for personalised respiratory health and pollution'' (EP/Y018680/1); INHALE, ``Health assessment across biological length scales'' (EP/T003189/1); AI4URBAN-HEALTH, ``AI Solutions to Urban Health Using a Place-Based Approach'' (UKRI~1241); D-XPERT, ``AI-Powered Total Building Management System'' (Innovate UK, TMF 10097909); the PREMIERE programme grant, ``AI to enhance manufacturing, energy, and healthcare'' (EP/T000414/1). The U-BIOPRED project has received funding from the Innovative Medicines Initiative (IMI) Joint Undertaking under grant agreement no.~115010, resources of which are composed of financial contributions from the European Union's Seventh Framework Programme (FP7/2007--2013), and European Federation of Pharmaceutical Industries and Associations (EFPIA) companies' in-kind contributions (www.imi.europa.eu). Support from Imperial-X's Eric and Wendy Schmidt Centre for AI in Science (a Schmidt Sciences programme) is also gratefully acknowledged.

\vspace{7pt}
\noindent\textbf{\textsf{Acknowledgements}}\\[5pt]
\small We would like to thank the following people who have kindly contributed their time and expertise to this paper: Dmytro Chupryna (Business Development and Strategic Partnerships Manager) and Daniil Mintc (Business Development and Community Manager) from OpenWeather Ltd.; Dan Hardman (Chief Technology Officer) from Coreblue Ltd.; and Yifeng Mao (AI Security and Privacy Lab) from Imperial College London.

\clearpage
\bibliographystyle{elsarticle-num-names}
\bibliography{refs}

\end{document}